\documentclass[10pt,twocolumn,letterpaper]{article}
\pdfoutput=1
\usepackage{iccv}
\makeatletter
\@namedef{ver@everyshi.sty}{}
\makeatother
\usepackage{tikz}
\usepackage{times}
\usepackage{epsfig}
\usepackage{graphicx}
\usepackage{amsmath}
\usepackage{amssymb}

\usepackage{xcolor}
\usepackage[normalem]{ulem}
\usepackage{svg}
\usepackage{float}
\usepackage{bbm}
\usepackage{bm}
\usepackage{multirow}

\usepackage[pagebackref=true,breaklinks=true,colorlinks,bookmarks=false]{hyperref}

\definecolor{dgreen}{rgb}{0.0,0.6,0.0} 
\definecolor{dred}{rgb}{0.6,0.0,0.0} 
\definecolor{BrickRed}{rgb}{0.72,0.0,0.0}%
\definecolor{grey}{rgb}{0.6,0.6,0.6}%

\newcommand{\pz}{\phantom{0}}
\newcommand{\empcell}{\makebox[0pt][l]{\phantom{0}--}\phantom{0.00}/\makebox[0pt][l]{\phantom{0}--}\phantom{0.00}}
\newcommand{\empcellnu}{\makebox[0pt][l]{\phantom{0}--}\phantom{0.00}}

\iccvfinalcopy 

\begin{document}

\title{On Exposing the Challenging Long Tail in Future Prediction of Traffic Actors}

\author{Osama Makansi \hspace{1cm} \"Ozg\"un \c{C}i\c{c}ek \hspace{1cm} Yassine Marrakchi \hspace{1cm} Thomas Brox\\
University of Freiburg\\
{\tt\small makansio,cicek,marrakch,brox@cs.uni-freiburg.de}
}

\maketitle

\begin{abstract}
Predicting the states of dynamic traffic actors into the future is important for autonomous systems to operate safely and efficiently. Remarkably, the most critical scenarios are much less frequent and more complex than the uncritical ones. Therefore, uncritical cases dominate the prediction. 
In this paper, we address specifically the challenging scenarios at the long tail of the dataset distribution. Our analysis shows that the common losses tend to place challenging cases sub-optimally in the embedding space. 
As a consequence, we propose to supplement the usual loss with a loss that places challenging cases closer to each other. This triggers sharing information among challenging cases and learning specific predictive features. 
We show on four public datasets that this leads to improved performance on the challenging scenarios while the overall performance stays stable.
The approach is agnostic w.r.t. the used network architecture, input modality or viewpoint, and can be integrated into existing solutions easily. Code is available at \href{https://github.com/lmb-freiburg/Contrastive-Future-Trajectory-Prediction}{github}.
\end{abstract}

\section{Introduction}
\begin{figure}[t]
\begin{center}
\includegraphics[width=1.0\columnwidth]{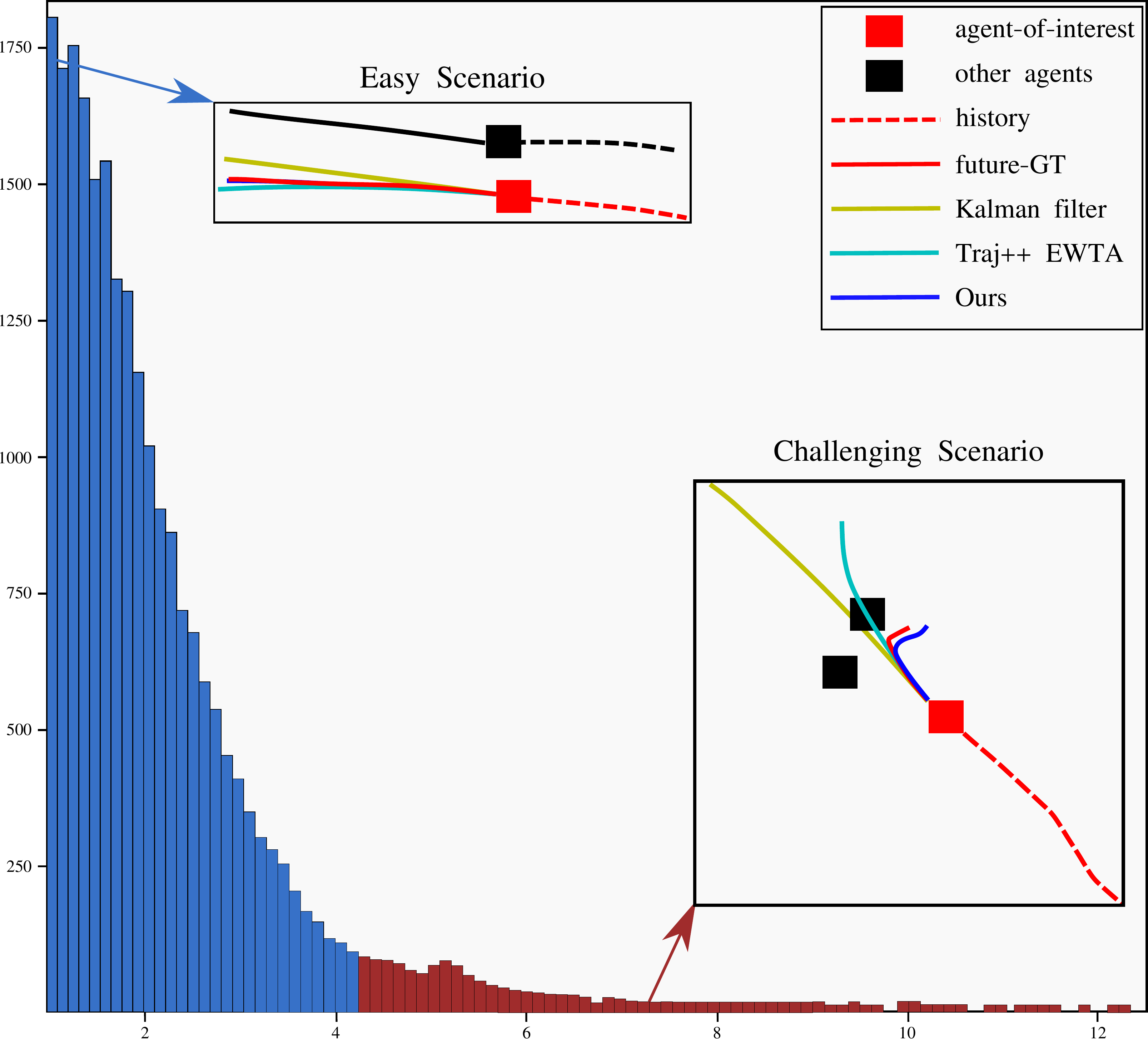}
\end{center}
   \vspace*{-4mm}
  \caption{Histogram of the ETH-UCY dataset based on the difficulty of the sample (based on displacement error of a Kalman filter \cite{Kalman1960}). An easy scenario (belongs to the head blue class) and a challenging scenario (belongs to the tail red class) are shown along the prediction of the state-of-the-art (Traj++ EWTA) and our approach. Our approach targets those challenging scenarios (from the tail) and improves their performance while maintaining a good performance on the easy scenarios.
  }
  \vspace*{-5mm}
\label{fig:teaser}
\end{figure}

Future prediction in traffic scenarios aims to foresee the future location of dynamic actors based on their current and previous locations and possibly other information about the environment. 
For an actor in interaction with others, reasoning about possible future locations of the other actors is necessary for path planning and to avoid collisions. Given enough data, some recent prediction methods~\cite{ewta,trajectron++,fln} also not just predict a single location of the actor in the future but a multimodal distribution over possible future locations. 

The average prediction errors of such methods look promising, but they hide that the training and test data is dominated by simple scenarios, where the trajectory can be smoothly propagated into the future. Such scenarios can be handled with a simple Kalman filter or other autoregressive models. However, the most safety-critical scenarios are those that involve close-by dynamic obstacles and require an evasive maneuver. Such scenarios are rare in both the training and the test data. The more complex and safety-critical they are, the less frequent they are. Fatal cases with a collision are not part of the dataset at all. 

As an example, the ETH-UCY dataset is often used to benchmark methods for future trajectory prediction. It is considered a challenging dataset, as it includes interacting pedestrians in crowded scenes. Figure~\ref{fig:teaser} shows the histogram of samples in this dataset based on their difficulty approximated by the prediction error of a Kalman filter. The large majority of scenarios can be well modeled by linear extrapolation, whereas scenarios that require more complex modeling are rare. The depicted challenging scenario showcases a pedestrian (red box), who will turn right in the future to avoid a collision with the stationary pedestrians (black boxes) in front of them. 

In this paper, we explicitly address the long-tailed data distribution in future prediction and focus on the rare but important cases rather than the average case. Straightforward ideas to re-distribute the dataset by undersampling the frequent scenarios \cite{undersample1,undersample2} or by reweighting the loss for these samples \cite{reweight4-inverse-effective} are not viable solutions, since it would reduce the (effective) data size dramatically. One can also oversample the challenging scenarios during training \cite{oversample4,oversample6-m2m}, yet this repetition of the same rare samples leads to overfitting and does not perform well, as we show in our experiments. Some works have tried to simulate rare cases \cite{forking,simaug}. However, to-date, even the most realistic simulations suffer from the domain gap between the simulated and the real world. An interesting direction for dealing with imbalanced data has been presented by Cao et al., who proposed a loss that ensures larger margins for the minority \cite{ldam-loss}.

We pick up this general idea and propose to reshape the feature embedding of the predictor. We show in a detailed analysis of the feature space that, with the usual loss, the challenging examples get placed next to many normal cases. Consequently, the relevant information of these samples gets smoothed out. As we push the challenging scenarios to be in proximity in the embedding, more of these samples that share a similar scenario build a small cluster and are no longer ignored. With this approach we can predict the future trajectory of interacting pedestrians better; see blue trajectory in Figure~\ref{fig:teaser}.

Our contributions can be briefly summarized as follows. (1) We analyze the problem of long-tailed data distributions in future prediction for the first time. (2) We propose a novel joint optimization of the regular regression loss for predicting the future location and a loss that reshapes the feature embedding in favor of the long-tail samples.
(3) We show that multi-headed networks outperform cVAEs in addressing the multimodal nature of the future. 

The proposed approach is easy to integrate into existing approaches, since it is agnostic to the network architecture, viewpoint, and input modalities. We demonstrate this by evaluating on four diverse public datasets. On each of them, the method improves the prediction quality of the challenging cases, while maintaining the quality on simple cases.

\section{Related Work}

\textbf{Future prediction.}
Deep learning methods dominate future prediction. LSTMs \cite{socialLSTM,CIDNN,srlstm,ContextAware,SceneLSTM,CarNet,csp} were mostly used to model the states of the agents over time, while graph-based approaches \cite{RSBG} were used to model the interactions between agents. However, these methods cannot handle the multimodal nature of the future.  
Meanwhile, several works addressed the multimodality in future prediction by cVAEs \cite{desire,pecnet}, GANs \cite{SocialGAN,Sophie,AgentTensor,socialBiGAT,reciprocal}, nonparametric approaches~\cite{forking,Relation} or a sampling-fitting framework~\cite{ewta}. 
Recently, graph neural networks \cite{stgcnn,trajectron++,evolvegraph,spagnn,lanegcn} and transformers \cite{star} have become popular to model the agent interactions. 
All aforementioned works assume that the scene is static and is observed from a bird's-eye view. Among these, Trajectron++ \cite{trajectron++} currently performs the best.

In automotive settings, the observation is typically from an egocentric view (e.g, with camera(s) or LiDAR mounted on the vehicle). This introduces new challenges due to the large egomotion of the vehicle and the narrow field of view. Multiple works project the data to the bird's-eye view using expensive 3D sensors \cite{Kinematic,Uber,Surround,Infer,TrafficPredict,Precog,Drogon}. 
Some recent approaches work directly on the egocentric view. Deterministic approaches \cite{Sted,Dtp} modeled the motion of the scene via optical flow. Yao et al. \cite{Ego} proposed to use the planned egomotion to improve the predictions. TraPHic~\cite{Traphic} exploited the interaction between nearby heterogeneous objects via LSTMs. Some works also tackled the multimodality in future prediction by using Bayesian RNNs to sample multiple futures with uncertainties \cite{Bayesian,Nemo}. Titan \cite{titan} modeled the future as a bi-variate Gaussian and conditioned the learning process on a set of labelled prior actions to further improve the prediction. Makansi et al.~\cite{fln} proposed a three-staged framework FLN-RPN, which currently performs the best in the egocentric view. 

None of the above approaches addressed the long tail of the data distribution. We base our method on Trajectron++ \cite{trajectron++} in the bird's-eye setting and FLN-RPN \cite{fln} for the egocentric setting, and specifically address the challenging cases in the long tail of the dataset distribution for the first time.






\textbf{Learning on imbalanced datasets.}
Issues with the long tail of a dataset have been well studied for classification tasks. Many works tackled the issue from the data side. A common approach is oversampling of rare classes~\cite{oversample0,oversample7,oversample2}. Another option is undersampling of the most frequent classes \cite{undersample1,undersample2}.
Several works follow the idea of generating more samples of the minority classes by simulation, which can be considered a more sophisticated version of oversampling~\cite{smote,oversample3,oversample5,oversample6-m2m}. Instead of changing the number of samples, samples can also be reweighted in the loss~\cite{reweight1-inverse-freq,reweight4-inverse-effective,reweight-by-hardness_0,reweight-by-hardness_1}. Some works proposed to learn these weights~\cite{reweight3,reweight5-meta-learning}. Recently, Li et al.~\cite{softmax-balanced} group classes of similar sizes and learn group-wise classifiers.

Another idea is to design loss functions that affect the feature space by increasing the inter-class distance and reducing the intra-class distance \cite{range-loss,reweight1-inverse-freq}. This concept of enlarging the margin between minority classes leads to a larger margin between classes and, thus, better generalization \cite{crl-loss,ldam-loss,metric-learning-uncertainty,affinity-loss}. Similarly, contrastive learning has become very popular due to promising results on self-supervised feature learning with noise-contrastive learning~\cite{nce,simclr,infonce}. Noisy versions of a sample (positives) are forced to be separated from other samples (negatives)~\cite{exemplarCNN}. Recently, contrastive learning enabled learning stronger feature extractors for classifying long-tail datasets \cite{Yang2020RethinkingTV}. 


All these methods were applied to classification tasks, where there is an explicit distinction between frequent and rare classes. Our approach also augments the loss to reshape the data distribution in the embedding space, yet we do not rely on predetermined clusters, since we have a regression task. Given the flexibility of contrastive learning in defining losses based on the definition of positive and negative samples, we adopt a novel way of embedding the samples based on their difficulty as measured by the performance of a Kalman filter and combine the reshaping of the embedding space with the regular regression loss. For sake of fair comparison, we also adapt previous methods tailored for classification and use them in conjunction with the regression loss as detailed in Section \ref{sec:exp}.

\begin{figure*}
\begin{center}
\includegraphics[width=1.0\textwidth]{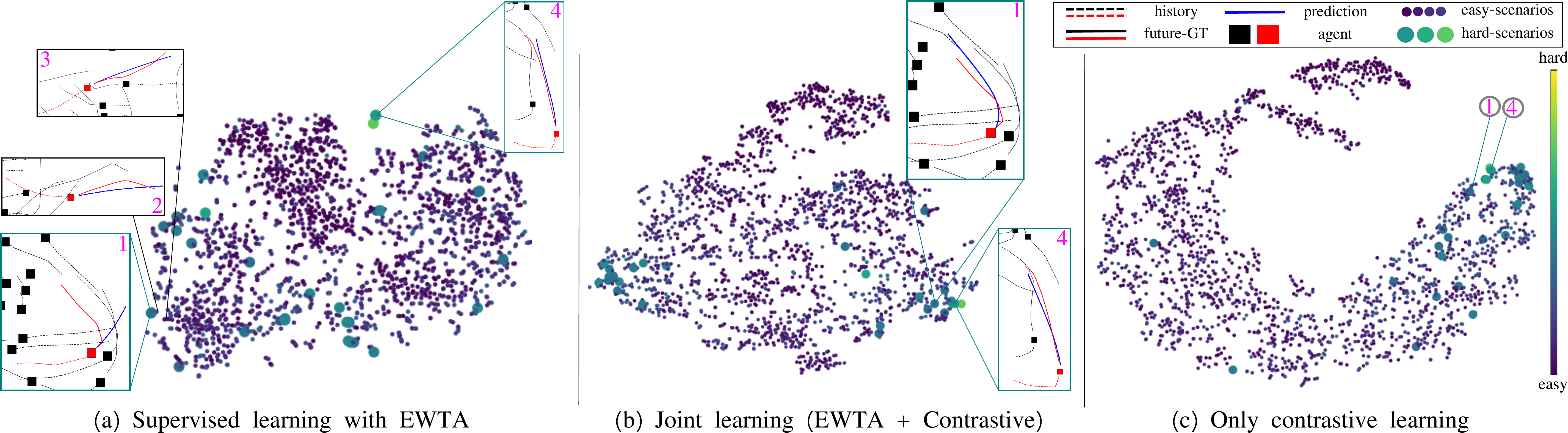}
\end{center}
\vspace*{-2mm}
  \caption{Feature space for the UNIV scene from ETH-UCY dataset using t-SNE \cite{tsne}. \textbf{(a)} Training only with the supervised objective for future prediction (e.g, EWTA). Rare challenging scenarios (large green bright circles) are scattered among the frequent easy scenarios (small dark blue circles). We zoom into two challenging (1,4) and two easy scenarios (2,3). \textbf{(b)} Joint learning with the supervised (EWTA) and the contrastive loss. The challenging scenarios form two sub-spaces where they can share relevant features. The two challenging examples (1,4) are close and benefit each other, which improves their future predictions considerably (particularly for 1).  \textbf{(c)} Only contrastive learning is used, where all challenging scenarios are strictly mapped to the same location. This destroys the task relevant cues and cannot provide any future prediction.}
  \vspace*{-2mm}
\label{fig:main}
\end{figure*}

\section{Future Prediction\label{sec:approach}}
Given current and past observations $(\mathbf{x}_{t-h},...,\mathbf{x}_{t})$, where $h$ is the length of the history, future prediction aims to predict the true state $\mathbf{y}$ of the actor of interest at times $(t + \Delta t,...,t + M\Delta t)$ in the future. An observation $\mathbf{x}$ at a single time step $t$ can consist of the 2D location $\mathbf{p}^{t}=(p_x,p_y)$, a map $\mathbf{Q}$ of the environment, a bounding box $\mathbf{b}^{t}=({b}_x,{b}_y,{b}_w,{b}_h)$ of the actor of interest, an RGB image $\mathbf{I}^{t}$, semantic segmentation $\mathbf{S}^{t}$, or future egomotion $\mathbf{e}^{t \Rightarrow t+\Delta t}$. For future trajectory prediction, the state $\mathbf{y}$ is defined as the future trajectory $(p_x,p_y)$ at $(t + \Delta t,...,t + M\Delta t)$ and for future localization prediction as the future bounding box $\mathbf{b}$ at $t + \Delta t$. 

We address the issue with the long tails of the data distribution in both bird's-eye view and egocentric settings. As backbone for these scenarios, we use the Trajectron++~\cite{trajectron++} and FLN-RPN~\cite{fln}, respectively. 

\subsection{Bird's-Eye View - Trajectron++}
Trajectron++~\cite{trajectron++} is the state-of-the-art method for future trajectory prediction in bird's-eye view. It takes the dynamic actors, the static environment, and heterogeneous input data into account. Given the past trajectories $[(p_x^{t-h},p_y^{t-h}),...,(p_x^{t},p_y^{t})]$, and optionally a map $\mathbf{Q}$ of the scene, Trajectron++ builds a directed spatiotemporal graph for a scene based on its topology. It predicts future trajectories $\mathbf{y}=[(p_x^{t + \Delta t},p_y^{t + \Delta t}),...,(p_x^{t + M\Delta t},p_y^{t + M\Delta t})]$. The nodes of the graph represent the actors, and the edges represent their interactions. The actors' histories are modeled by LSTMs, features of interacting actors are aggregated via point-wise summation, and GRUs are used to decode the future trajectories. The original architecture employs a cVAE to produce multiple future trajectories. 

Since cVAEs require multiple runs of the decoder to obtain multiple predictions, we replace the cVAE by the multi-hypotheses networks trained with EWTA (Evolving Winner-Takes-All)~\cite{ewta}. The EWTA loss for every sample $i$ in the batch is defined as: 
\begin{eqnarray}
    L_{i}^{\rm EWTA}  &=& \sum_{m=1}^{M} \sum_{k=1}^{K} w_{k,m} ||\mathbf{p}^{t+m\Delta t}_k - \mathbf{\hat{p}}^{t+m\Delta t}||
    \label{eq:ewta1}
        \mathrm{\,,}\\  
    w_{j,m} &=& \mathbbm{1}_{j \in \underset{k}{\mathrm{argmin}}\, ||\mathbf{p}^{t+m\Delta t}_k - \mathbf{\hat{p}}^{t+m\Delta t}||}
    \mathrm{\,,}
    \label{eq:ewta2}
\end{eqnarray}
where $K$ is the number of estimated hypotheses. $\mathbbm{1}_{cond}$ is the indicator function that returns $1$ if the condition $cond$ returns true and $0$ otherwise. $\mathbf{p}^{t+m\Delta t}_k$ and $\mathbf{\hat{p}}^{t+m\Delta t}$ denote the $k$th predicted future state and the ground truth at future time step ($t+m\Delta t$), respectively. The $\mathrm{argmin}$ returns the $k$ hypotheses closest to the ground truth, where $k$ gradually decreases from $K$ to $1$ during training. While all hypotheses are penalized in the beginning of the training, only the best one would be penalized at the end of the training.
The Trajectron++ augmented by EWTA (Figure~\ref{fig:overview} (top)) produces multiple future trajectories in a single network pass and outperforms the standard Trajectron++, as we show in Section~\ref{sec:exp}.

\subsection{Egocentric View - FLN-RPN}
FLN-RPN~\cite{fln} is the state-of-the-art method for future localization prediction in the egocentric setting. FLN-RPN predicts the multimodal distribution of the future localization of an actor in three steps. First, it predicts where an actor is most likely to be in the current image (\emph{Reachability Prior}). Second, it transfers the reachability prior from the current frame to the future frame using the future egomotion. Finally, past bounding boxes of the actor $\mathbf{b}$, images $\mathbf{I}$, semantic segmentations $\mathbf{S}$ at time steps ($t-h,...,t$), the future egomotion $\mathbf{e}^{t \Rightarrow t+\Delta t}$ and the predicted future reachability prior are given to the network to predict the future localization of the actor of interest. The prediction has the form of set of bounding boxes $\mathbf{b}$ at $t + \Delta t$. The two key components of FLN-RPN are the reachability prior, which helps overcoming mode collapse, and the EWTA loss function (Eq.~\ref{eq:ewta1} with $M=1$ and $\mathbf{p}$ is replaced by $\mathbf{b}$) that can learn diverse multiple states of the future in a single forward pass. Figure~\ref{fig:overview} (bottom) illustrates the FLN-RPN framework. 

\subsection{Difficulty Ranking\label{sec:difficulty}}
Before we explore the effects of the distribution of the challenging scenarios in the feature space on the final prediction, we need to know how challenging each scenario is. Since manual labeling is not a viable option, we use a common and simple metric to measure the difficulty of cases: the displacement error made by the Kalman Filter~\cite{Ego,fln,Kalman1960} on this sample. Small errors indicate good approximation with linear extrapolation, whereas large errors indicate a challenging scenario that requires complex nonlinear prediction. 

\section{Why are Hard Cases Ignored by the Model?}
To understand the cause of the problem with samples from the long tail of the data distribution, we visualized the feature embedding of the data from a network trained with a supervised future prediction objective and analyzed particular cases in detail.  Figure~\ref{fig:main} (left) shows the feature space for the UNIV scene in the ETH-UCY dataset projected to 2D with t-SNE \cite{tsne}. Each dot is a sample from the scene mapped to the feature space by the network trained with EWTA loss (Eq.~\ref{eq:ewta1}). Hard cases are sprinkled among the easy cases in the feature space without any structure. A closer look at a hard case (1) reveals that it shares some similarity with corresponding easy cases (2,3), which explains its position in the embedding, but the relevant cues, in which it is different from the easy cases, get ignored with normal training. The sample should rather be close to another challenging example (4) to capture the social interaction, where pedestrians walk in groups and follow other groups. 
We believe that challenging scenarios being alone in a manifold full of easy scenarios causes the network to ignore them and base its decisions on shortcuts learned from the dominant easy scenarios. The network does not get a chance to learn dedicated features to solve challenging cases by reusing some common cues among them (1,4), as long as they get mixed up with the easy cases. Indeed, the prediction for case (1) is quite wrong since it is similar to the prediction of cases (2,3), where social interaction is missing.


\section{Reshaping the Embedding with Contrastive Learning}

\begin{figure}[t]
\includegraphics[width=1.0\linewidth]{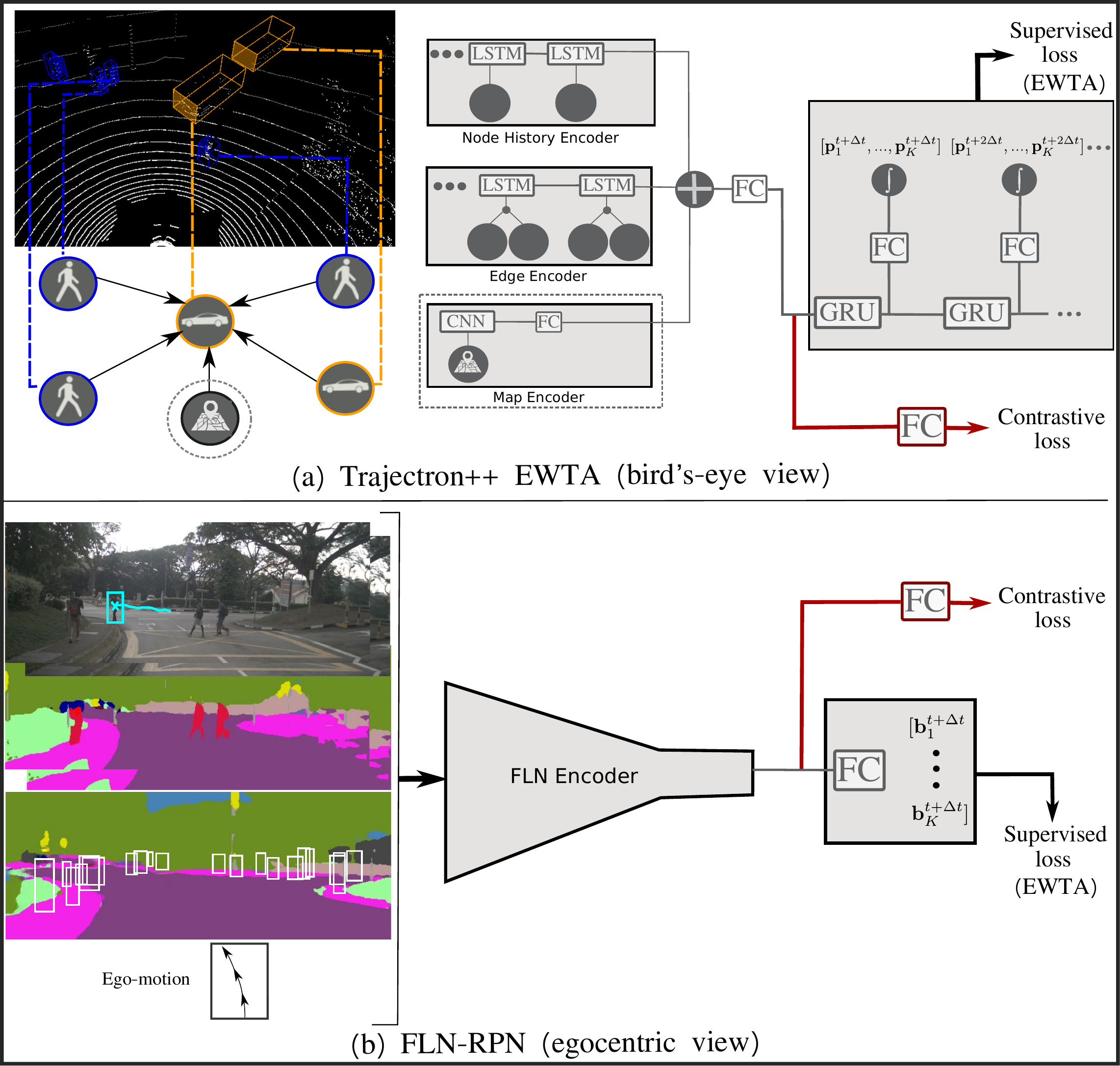}
\vspace*{-4mm}
  \caption{Schematic that shows how we flexibly integrate the contrastive loss (red) in existing future prediction frameworks. (a) Bird's-eye view (Trajectron++\cite{trajectron++}). (b) Egocentric view (FLN-RPN\cite{fln}). 
  Independent of the contrastive loss, we modified Trajectron++ by replacing the cVAE with the EWTA~\cite{ewta} framework to better capture the multimodality of the predicted future and for faster inference time. The map encoder (dashed gray) is optional and only used for the nuScenes dataset.}
\label{fig:overview}
\vspace*{-3mm}
\end{figure}

The analysis from the previous section triggers the idea to push hard samples away from the easy ones, such that the relevant cues of similar hard samples get the chance to be no longer ignored during training. We implement this idea with an additional contrastive loss. 
Contrastive learning enforces certain training samples (positives) to be closer in the embedding to a sample (anchor) $i$ than others (negatives). There are multiple ways to express this in a loss. The most popular is 
\begin{equation} \label{eq:cont}
\begin{split}
L_{i}^{\rm Contr} & = -\frac{1}{N_{\mathbf{po}_i}-1}\sum_{j=1}^{N}{\mathbbm{1}_{i\neq{j}}}\cdot{\mathbbm{1}_{j \in \mathbf{po}_{i}}} \\
 &  \cdot{\log\frac{\exp(\mathbf{z}_{i}\boldsymbol{\cdot}{\mathbf{z}_{j}}/\tau)}{\sum_{k=1}^{N}{\mathbbm{1}_{i\neq{k}}\cdot{\exp(\mathbf{z}_{i}\boldsymbol{\cdot}{\mathbf{z}_{k}}/\tau)}}}} \mathrm{\,,}
\end{split}
\end{equation}
where $z$ is the learned feature vector at the bottleneck of the network (see Figure~\ref{fig:overview}), $\mathbf{po}_{i}$ is the positive set of anchor $i$. $\mathbbm{1}_{cond}$ is the indicator function that returns $1$ if the condition $cond$ returns true and $0$ otherwise. $N$ is the total number of samples in the batch. $N_{\mathbf{po}_{i}}$ is the total number of positive samples for the anchor $i$. $\tau > 0$ is the temperature parameter. Positive samples are often defined as augmented versions of the same image \cite{simclr} or samples belonging to the same class \cite{sup-con}. Negative samples, on the other hand, are other samples in the batch that do not satisfy the positive criterion by some safe margin. Since our goal is to distribute the features based on the difficulty, we define the positive set $\mathbf{po}_{i}$ as the set of samples $j$ in the batch which has a difficulty score $s_j$ satisfying $|s_{i}-s_{j}|<\theta_{p}$, where $\theta_{p}$ is a hyper-parameter defining the positivity threshold. Similarly, the negatives samples are defined as all samples with a difficulty score satisfying $|s_{i}-s_{j}|>\theta_{n}$. Note that we use different thresholds $\theta_{p} \neq \theta_{n}$ implying that many samples in the batch are neither positive nor negative. In order to minimize this loss, the network must maximize the nominator and minimize the denominator. Doing so, it learns to map the positive samples close in the feature space and the negative ones apart. The result of training with such a loss is shown in Figure~\ref{fig:main} (right).

While having the hard cases being pushed together is good for them to share relevant cues and learn prediction models for less common scenarios, there is much diversity among hard cases, and not all of them should be pushed to the same space. In particular, we should not destroy cues shared with the easy examples, which are necessary for the network to solve the actual task. The contrastive loss alone can not predict the future state. 
To this end, we jointly optimize for the supervised future prediction loss $L^{\rm EWTA}$ and the self-supervised contrastive loss $L^{\rm Contr}$ as: 
\begin{equation}
    L_{i} = L_{i}^{\rm EWTA} + \lambda \cdot L_{i}^{\rm Contr},
\end{equation}
where $\lambda$ controls the importance of the contrastive loss, hence the strength of the attraction that pulls hard cases together.

Figure~\ref{fig:main} (middle) shows the effect of this combination. 
Cases (1) and (4) fall into the same sub-space resulting in a much better prediction for (1). Other hard cases rather stay with similar easy samples as they have no other hard cases to share information with. 

Due to its simplicity, this difficulty-based contrastive learning can be added to any existing method as long as the difficulty can be defined explicitly on the training set.

\section{Experiments\label{sec:experiments}}

\subsection{Datasets}

The \textbf{ETH-UCY} dataset is the combination of the ETH~\cite{eth} and the UCY~\cite{ucy} pedestrian datasets. Both include videos from bird's-eye view of the pedestrians, where the trajectories are manually annotated. The challenges in these datasets are the frequent interactions between pedestrians, as the scenes are very crowded, and the lack of visual information due to the viewpoint, i.e, the actors are small and uninformative. We present 5-fold cross-validation results on the five scenes of the dataset.

\textbf{nuScenes}~\cite{nuscenes} is a large autonomous driving dataset with 1000 scenes, where each is 20 seconds long. It provides HD semantic maps with 11 different layers and accurate bounding box annotations in time. It provides scenarios from bird's-eye view and egocentric view, and we experiment with each of them.

\textbf{Waymo}~\cite{waymo} is the most recent autonomous driving dataset with 1000 scenes, where each is 20 seconds long. We use the validation part of the dataset (202 scenes) to show zero-shot transfer of our approach in egocentric view (i.e, without retraining the model).

\subsection{Evaluation Metrics}

\textbf{min-ADE} is the minimum average displacement error. It computes the mean $L_{2}$ distance between all predicted trajectories and the ground truth and reports the error of the closest one. This is sometimes also referred to as \emph{oracle} (or \emph{best-of-many}), since the selection of the minimum error depends on the ground truth.

\textbf{min-FDE} is the minimum final displacement error. It computes the $L_{2}$ distance between the final locations of the predicted trajectories and the ground truth at the end of the predicted time horizon ($t + M\Delta t$) and, like min-ADE, reports the minimum.


\subsection{Training Details}
In our experiments for bird's-eye view, we followed the original training schedule for Trajectron++ \cite{trajectron++}. We trained the Trajectron++ (EWTA) with batch size 256 for 100 epochs in every EWTA stage ($k=K,...,1$) for ETH-UCY and for 5 epochs in every EWTA stage for nuScenes. For the experiments in egocentric view, we used ResNet34 \cite{resnet} as the encoder of FLN-RPN \cite{fln} and trained with batch size of 32. Following \cite{trajectron++,fln}, we set $M$ to 12, 6, 1 and $\Delta t$ to 0.4, 0.5, 3.0 for ETH-UCY, nuScenes (bird's-eye view) and nuScenes/Waymo (egocentric view), respectively. The remaining design choices were kept as in the original papers \cite{trajectron++,fln}. For our joint optimization, $\lambda$ was chosen based on the validation set as 1, 50, 150 for nuScenes (bird's-eye view), ETH-UCY, and nuScenes (egocentric view), respectively. We used the recommended value of $0.5$ for $\tau$ \cite{simclr}. $\theta_{p}$ and $\theta_{n}$ were set such that the ratio of positives and negatives over the batch size are 10\% and 40\% for Trajectron++ and 33\% and 33\% for FLN-RPN, respectively. $z$ had the dimensions of 232 and 256 for Trajectron++ and FLN-RPN, respectively. A study about the effect of the hyper-parameter $\lambda$ is presented in the supplemental material.

\subsection{Baselines}
\textbf{Bird's-eye view (ETH-UCY).} We selected a set of recent methods addressing the future trajectory prediction: Graph-based approaches: RSBG\cite{RSBG}, S-STGCNN\cite{stgcnn}, and Trajectron++\cite{trajectron++} (referred as Traj++); transformer-based approach: STAR\cite{star}; multi-stage networks: TPNet \cite{tpnet} and PECNet \cite{pecnet}.

\textbf{Bird's-eye view (nuScenes).} We compare against a set of baselines including deterministic LSTM-based approaches: S-LSTM \cite{socialLSTM}, CSP \cite{csp}, and CAR-Net \cite{CarNet}; multimodal graph-based approaches: SpAGNN \cite{spagnn} and Trajectron++ \cite{trajectron++}.

\textbf{Egocentric view.} We compare against the multimodal state-of-the-art FLN-RPN \cite{fln}. 

Moreover, for all settings and datasets, we implemented the common approaches for imbalanced data: resampling~\cite{oversample0}, reweighting using the inverse class frequency~\cite{reweight1-inverse-freq}, and reweighting using the effective number of samples~\cite{reweight4-inverse-effective}. We also adapt sophisticated long-tail classification methods\cite{ldam-loss,softmax-balanced} to the considered task by defining classes based on the discretization of Kalman filter scores. Then, the network is jointly trained on the regression loss and the considered classification loss (more details are provided in the supplementary). Notice that recent methods: cRT, $\tau$-norm and LWS introduced by Kang et al.~\cite{sota-class-long-tailed} can not be adapted to regression tasks since they do not affect the feature extractor and fully rely on post-processing the classifier which is not needed at test time in our scenario.

\subsection{Results \& Discussion \label{sec:exp}}
To show the validity of the proposed approach, we selected strong baselines and state-of-the-art methods for comparison.  Tables~\ref{tab:eth}, \ref{tab:nuScenes_top} and \ref{tab:ego} summarize our results on the four different datasets. Since we are interested in improving the quality of the predictions of the rare cases, we report min-ADE and min-FDE for all samples, as well as the top 1-3\% challenging cases.

\textbf{EWTA vs cVAE.} Tables \ref{tab:eth} and \ref{tab:nuScenes_top} show that our base method, where we use the Trajectron++ as the backbone with the EWTA objective, clearly outperforms the previous state-of-the-art Trajectron++. This shows that EWTA-based sampling for possible future trajectories works better than cVAE-based sampling.

\begin{table}
\centering
\resizebox{1.0\linewidth}{!}{%
\begin{tabular}{|l|c||c|c|c|}%
\hline
                                    & All                & Top 3\%              & Top 2\%              & Top 1\% \\
\hline
RSBG \cite{RSBG}                    & 0.48/0.99          &   -/-                & -/-                  & -/-      \\
Reciprocal \cite{reciprocal}        & 0.44/0.90          &   -/-                & -/-                  & -/-      \\
TPNet \cite{tpnet}                  & 0.42/0.90          &   -/-                & -/-                  & -/-      \\
S-STGCNN \cite{stgcnn}              & 0.44/0.75          &   -/-                & -/-                  & -/-      \\
STAR \cite{star}                    & 0.26/0.53          &   -/-                & -/-                  & -/-      \\
PEC-NET \cite{pecnet}               & 0.29/0.48          &   -/-                & -/-                  & -/-      \\
Traj++ \cite{trajectron++}          & 0.21/0.41          &   0.65/1.42          & 0.71/1.51            & 0.58/1.23 \\
\hline
Traj++ EWTA (ours)                  & 0.16/0.32          &   0.47/1.07          & 0.51/1.13            & 0.42/0.87 \\
\hline
\hline
+ LDAM~\cite{ldam-loss}             & 0.17/0.33          & 0.47/1.04            & 0.50/1.08            & 0.42/0.83\\
+ LDAM-DRW~\cite{ldam-loss}         & 0.17/0.33          & 0.47/1.04            & 0.51/1.08            & 0.43/0.83\\
+ BAGS~\cite{softmax-balanced}      & 0.17/0.32          & 0.48/1.08            & 0.51/1.10            & 0.42/0.85\\
\hline
+ contrastive (ours)                & \textbf{0.16/0.32} & \textbf{0.46/1.03}   & \textbf{0.48/1.03}   & \textbf{0.38/0.71} \\
\hline
\end{tabular}
}
\vspace*{-1mm}
    \caption{Average error on the \textbf{ETH-UCY benchmark} over all test samples and over the 1-3\% most challenging scenarios in the format of (min-ADE/min-FDE). Joint learning with the contrastive loss yields large improvements on the challenging scenarios while not harming the overall average accuracy.}
    \label{tab:eth} 
    \vspace*{-3mm}
\end{table}

\begin{table}[ht!]
\centering
\resizebox{1.0\linewidth}{!}{%
\begin{tabular}{|l|c||c|c|c|}%
\hline
                                & All                & Top 3\%            & Top 2\%            & Top 1\%\\
\hline
S-LSTM \cite{socialLSTM}        & \empcellnu/1.61    & \empcell           & \empcell           & \empcell  \\
CSP \cite{csp}                  & \empcellnu/1.50    & \empcell           & \empcell           & \empcell  \\
CAR-Net \cite{CarNet}           & \empcellnu/1.35    & \empcell           & \empcell           & \empcell  \\
SpAGNN \cite{spagnn}            & \empcellnu/1.23    & \empcell           & \empcell           & \empcell  \\
Traj++ \cite{trajectron++}      & 0.22/0.39          & 0.55/0.98          & 0.60/1.04          & 0.72/1.21\\
\hline
Traj++ EWTA (ours)              & 0.19/0.32          & 0.48/0.88          & 0.50/0.88          & 0.59/1.02\\
\hline
\hline
+ LDAM~\cite{ldam-loss}        & 0.18/0.32          & 0.48/0.88          & 0.51/0.93          & 0.60/1.10\\
+ LDAM-DRW~\cite{ldam-loss}    & 0.18/0.32          & 0.50/0.93          & 0.52/0.96          & 0.63/1.14\\
+ BAGS~\cite{softmax-balanced} & 0.18/0.31          & 0.48/0.88          & 0.51/0.94          & 0.61/1.11\\
\hline
+ contrastive (ours)           & \textbf{0.18/0.30} & \textbf{0.44/0.73} & \textbf{0.46/0.72} & \textbf{0.54/0.85}\\
\hline
\end{tabular}
}
\vspace*{-1mm}
    \caption{Average error on the \textbf{nuScenes dataset (bird's eye view)} over all test samples and over the 1-3\% most challenging scenarios in the format of (min-ADE/min-FDE). Joint learning with the contrastive loss yields large improvements on the challenging scenarios and even improves the overall average accuracy a little.}
    \label{tab:nuScenes_top} 
    \vspace*{-3mm}
\end{table}

\begin{table*}[t]
\centering
\resizebox{0.7\textwidth}{!}{%
\begin{tabular}{|l|c|c|c|c||c|c|c|c|}%
\hline
                               & \multicolumn{4}{c||}{nuScenes Egocentric View}   & \multicolumn{4}{c|}{Waymo Egocentric View}\\
\hline
                               & All   & Top 3\% & Top 2\% & Top 1\%              & All           & Top 3\% & Top 2\% & Top 1\% \\
\hline
FLN-RPN \cite{fln}             & 7.10  & 29.98   & 31.13   & 36.16                & \textbf{6.39} & 24.87   & 25.49   & 27.32    \\
\hline
+ LDAM~\cite{ldam-loss}        & 8.04  & 25.23   & 26.02   & 31.13                & 7.61          & 23.00   & 23.09   & 25.05      \\
+ LDAM-DRW~\cite{ldam-loss}    & 8.01  & 26.63   & 27.85   & 34.58                & 8.05          & 25.23   & 25.98   & 29.32      \\
+ BAGS~\cite{softmax-balanced} & 7.28  & 29.54   & 30.38   & 35.74                & 6.67          & 24.45   & 24.88   & 26.66      \\
\hline
+ contrastive (ours)           & \textbf{7.04} &\textbf{25.05} & \textbf{25.26} & \textbf{27.49}  & 6.49 & \textbf{22.36} & \textbf{22.72} & \textbf{24.09}     \\
\hline
\end{tabular}
}
\vspace*{1mm}
    \caption{Results on \textbf{egocentric datasets (nuScenes and Waymo)}. We show the min-FDE over all scenarios and over the top 1-3\% challenging scenarios. Our approach yields an improvement on the challenging scenarios while maintaining the performance on average.}
    \label{tab:ego} 
    \vspace*{-5mm}
\end{table*}

\textbf{Large improvements on the challenging cases.} Results on all datasets show that our approach yields large improvements on the challenging cases (particularly for the top 1\%) while maintaining the overall average error. In particular, on the most challenging cases (top 1\%), our approach improves by 18\%, 17\%, 23\% and 12\% on the ETH-UCY, nuScenes (bird's eye-view), nuScenes (egocentric view) and Waymo open dataset, respectively. The challenging training samples, as hypothesized, help each other when they are in proximity in the feature space. Notably, the studied datasets differ in their input modalities (additional semantic maps for nuScenes), viewpoint (bird's-eye vs egocentric views), and prediction output (2D points in bird's-eye view while bounding boxes for egocentric view). This indicates that the approach is agnostic to different input modalities and generalizes well.

\textbf{Comparison to long-tail classification baselines.} Tables~\ref{tab:eth}, \ref{tab:nuScenes_top} and \ref{tab:ego} show a comparison against recent methods addressing the long-tail problem in  classification. Our method based on the contrastive loss outperforms all these techniques on all metrics. 

\textbf{Zero-shot transfer.} Results on the Waymo dataset (Tab.~\ref{tab:ego}) show promising zero-shot transfer to unseen dataset, where models were trained on the nuScenes dataset and tested on the validation split of the Waymo dataset.  

\textbf{Avoids bias.} In Table~\ref{tab:overfitting} we compare our method against the common approaches for imbalanced data: resampling and reweighting. We report across all datasets the performance over all samples and over the most challenging samples (top 1\%). As expected, these baselines tend to bias the challenging cases. Hence, the average performance drops significantly (66\%, 16\%, 44\% and 64\% for ETH-UCY, nuScenes bird's, nuScenes egocentric and Waymo). Our method, on the other hand, maintains the average performance over all samples. Detailed results on all metrics and difficulties are provided in the supplementary.

\begin{table}[hb!]
\centering
\resizebox{1.0\linewidth}{!}{%
\begin{tabular}{|l|c|c|c|c|}%
\hline
                                              & ETH-UCY            & nuScenes-B          & nuScenes-E           & Waymo \\
                                              & All/Top 1\%        & All/Top 1\%         & All/Top 1\%          & All/Top 1\% \\
\hline
Baseline                                      & 0.32/0.87          &  0.32/1.02          & \pz7.10/36.16        & \pz\textbf{6.39}/27.32       \\
\hline
+ resample~\cite{oversample0}                 & 0.53/1.22          &  0.37/1.33          & 10.20/21.62          & 10.48/19.69       \\
+ reweight~\cite{reweight1-inverse-freq}      & 0.56/0.76          &  0.58/1.67          & 14.47/16.20          & 14.00/\textbf{16.44} \\
+ reweight~\cite{reweight4-inverse-effective} & 0.56/0.78          &  0.60/1.71          & 16.54/\textbf{15.46} & 17.43/18.79       \\
\hline
+ contrastive                                 & \textbf{0.32/0.71} &  \textbf{0.30/0.85} & \pz\textbf{7.04}/27.49  & \pz6.49/24.09       \\
\hline
\end{tabular}
}
\vspace*{0mm}
    \caption{Comparison to the common resampling/reweighting techniques on the four datasets. For each method, we show the min-FDE over all samples and over top 1\% challenging samples. Our method yields large improvements on the challenging ones while maintaining the average. This is in contrast to the reweighting/resampling baselines, which lead to much worse performance on average. Baseline indicates Traj++ EWTA for bird's eye view and FLN-RPN for egocentric view.}
    \label{tab:overfitting} 
    \vspace*{-5mm}
\end{table}

\subsection{Qualitative Results}
\begin{figure*}[t!]
\begin{center}
\includegraphics[width=0.95\textwidth]{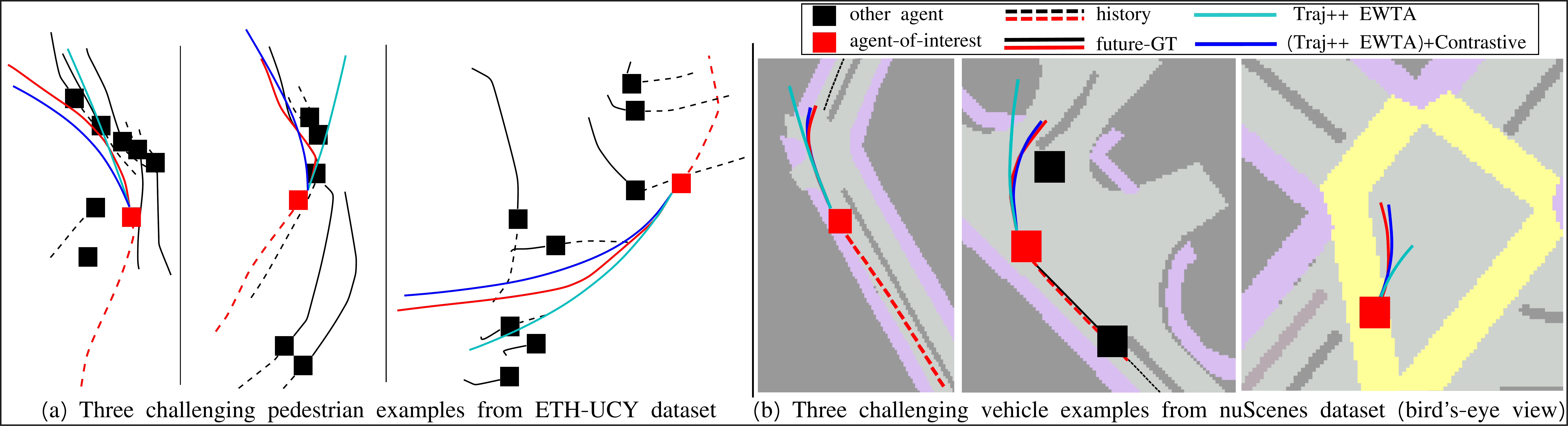}
\end{center}
\vspace*{-4mm}
  \caption{Qualitative challenging examples for pedestrians from ETH-UCY dataset (a) and vehicles from nuScenes bird's-eye view dataset (b). Note how \textcolor{blue}{our approach} outperforms the SOTA \textcolor{cyan}{(Trajectron++ EWTA)} by generating a future trajectory closer to the \textcolor{red}{ground truth}. We visualize the best hypothesis for each method. For the examples from nuScenes (b), we show the underlying map on which the method need to reason about.}
\label{fig:ex1}
\vspace*{-1mm}
\end{figure*}

In Figure~\ref{fig:ex1} (a), we show three challenging examples from ETH-UCY dataset. In all the cases, the future trajectory of the pedestrian (red) is not trivial, and the network must model the interaction between pedestrians to generate a plausible future trajectory. 
Our approach (blue) generates trajectories that are much closer to the ground truth than Trajectron++ EWTA (cyan). In Figure~\ref{fig:ex1} (b), we show three challenging examples for vehicles from the nuScenes dataset (bird's-eye view). In these examples, the vehicle changes direction, which requires interpretation of the maps. Our approach succeeds on these examples, whereas Trajectron++ EWTA misses these cues and predicts the simple continuation of the trajectory. 

\begin{figure*}[t!]
\begin{center}
\includegraphics[width=0.95\textwidth]{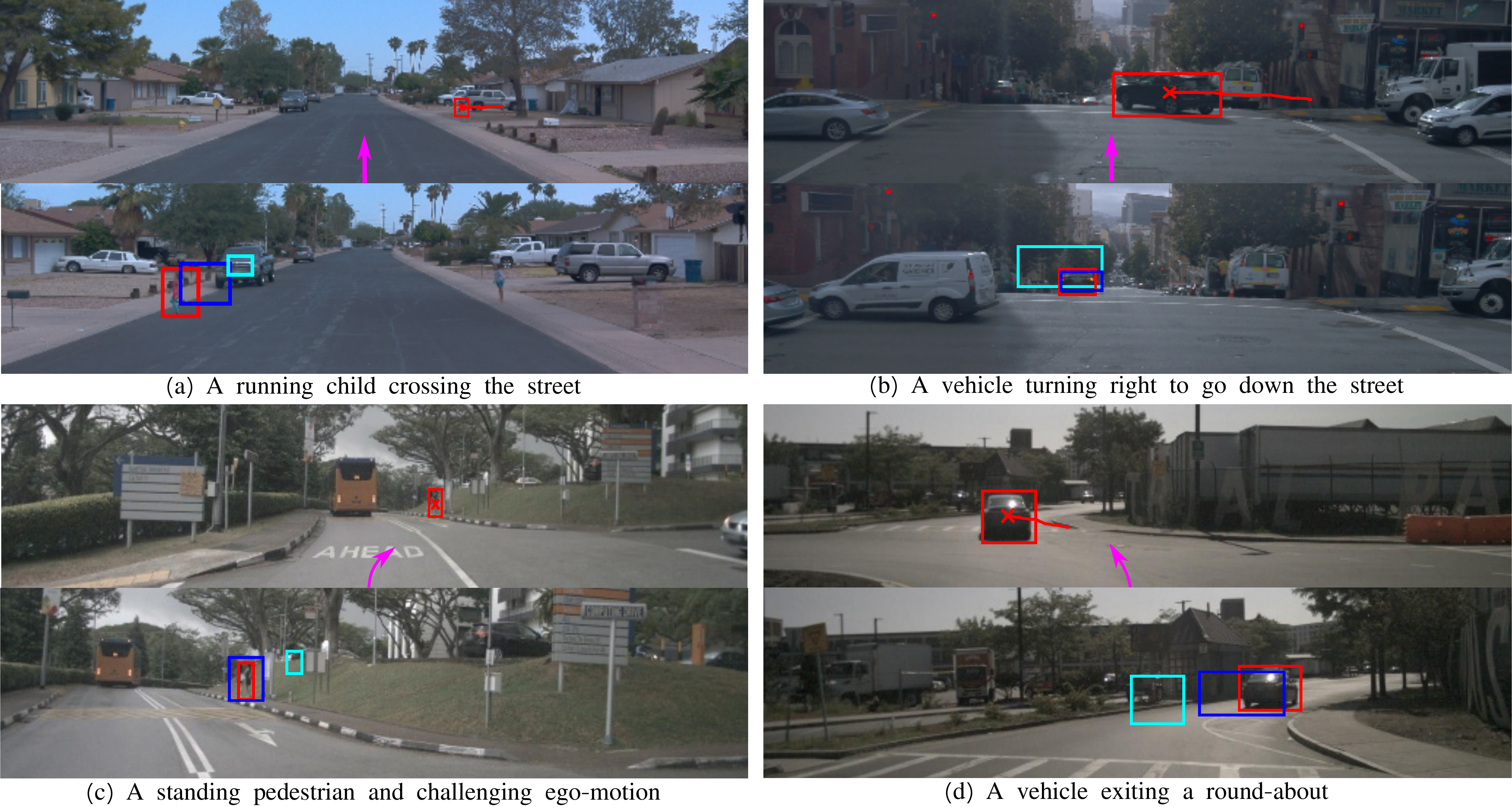}
\end{center}
\vspace*{-4mm}
  \caption{Qualitative challenging examples from Waymo open dataset (a-b) and nuScenes egocentric view (c-d). For each example, we show both the last observed image (top) and the future image (bottom) along with the predictions (\textcolor{cyan}{FLN-RPN \cite{fln}} and \textcolor{blue}{Ours}) and the \textcolor{red}{ground truth}. We visualize the best hypothesis for each method. The \textcolor{magenta}{future egomotion} is also shown as arrow indicating the motion of the ego-car.}
\label{fig:ex2}
\vspace*{-1mm}
\end{figure*}

Figure~\ref{fig:ex2} shows four different examples from the egocentric setting. Figure~\ref{fig:ex2} (a) shows a child crossing the street in front of the vehicle. Figure~\ref{fig:ex2} (b) shows a vehicle that will turn right to go down the street, which is rarely encountered. Figure~\ref{fig:ex2} (c) shows an example that is difficult because of the uncommon egomotion of the car moving to the opposite lane to overtake the bus. Figure~\ref{fig:ex2} (d) shows a vehicle that turns right to exit the round-about. In all these examples, our approach makes predictions close to the ground truth (both in scale and location), whereas the baseline fails.

\begin{figure*}[t!]
\begin{center}
\includegraphics[width=0.95\textwidth]{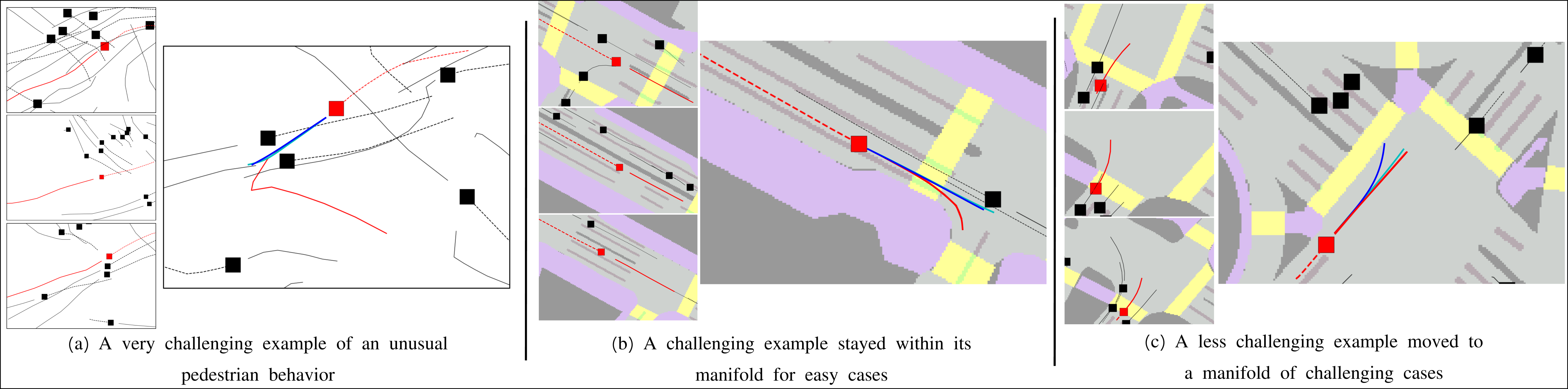}
\end{center}
\vspace*{-4mm}
  \caption{Three examples for different categories of failures from our method. Each example is shown together with three other examples
on its left from its manifold resulting from our approach. (a) An example of a pedestrian from ETH-UCY dataset who unexpectedly
decided to turn back and go left. Such an unexpected future behavior is very hard to model. (b) A vehicle from nuScenes (top view) dataset
that decided to turn right and our approach in unable to change its manifold. (c) An example of a less challenging example from nuScenes
(bird’s-eye view) dataset which our approach mistakenly moves to a challenging manifold.}
\label{fig:failure}
\vspace*{-10mm}
\end{figure*}

\textbf{Limitations and failure cases.} We also analyzed the limits  of our approach to identify room for further improvements. We found that some challenging cases continue to stay in a manifold for easy cases because of missing similarity to other hard cases Figure~\ref{fig:failure} (b), or easy cases moved wrongly to a manifold of challenging cases Figure~\ref{fig:failure} (c). Consequently, our approach yields wrong prediction. We also found that our method, like other methods, cannot model unexpected behavior, such as suddenly stopping and turning in the opposite direction Figure~\ref{fig:failure} (a). We also provide the feature embedding before and after application of our approach for all datasets in the supplementary. 

\section{Conclusions}

We addressed the long-tailed data distributions by acting on the feature embedding. We showed that pulling the rare challenging samples together in the feature embedding via contrastive learning helps improve their final predictions while preserving the performance over the whole dataset. We validated our approach qualitatively and quantitatively on four different datasets, two different viewpoints and different combinations of input and output modalities. The proposed loss can be integrated easily into existing approaches to improve their performance on critical challenging cases. We hypothesize that the concept is generic and could be integrated into other regression tasks with an unbalanced sample distribution, as long as there is a way to identify the underrepresented samples during training.

\section{Acknowledgments}
The research leading to these results was funded by the German Federal Ministry for Economic Affairs and Energy within the project “KI Delta Learning" (19A19013N) and by the Excellence Strategy of the German Federal and State Governments, (CIBSS - EXC 2189).

\vfill
{\small
\bibliographystyle{ieee_fullname}
\bibliography{arxiv.bib}
}

\cleardoublepage

\twocolumn[
\null
\vskip .375in
\begin{center}
  {\Large \bf Supplementary Material for: \\ 
  On Exposing the Challenging Long Tail in Future Prediction of Traffic Actors \par}
  \vspace*{24pt}
  {
  \large
  \lineskip .5em
  \par
  }
  \vskip .5em
  \vspace*{12pt}
\end{center}]

\setcounter{section}{0} 

\maketitle
\section{Visualization Plots}
\begin{figure*}
\begin{center}
\includegraphics[width=1.0\textwidth]{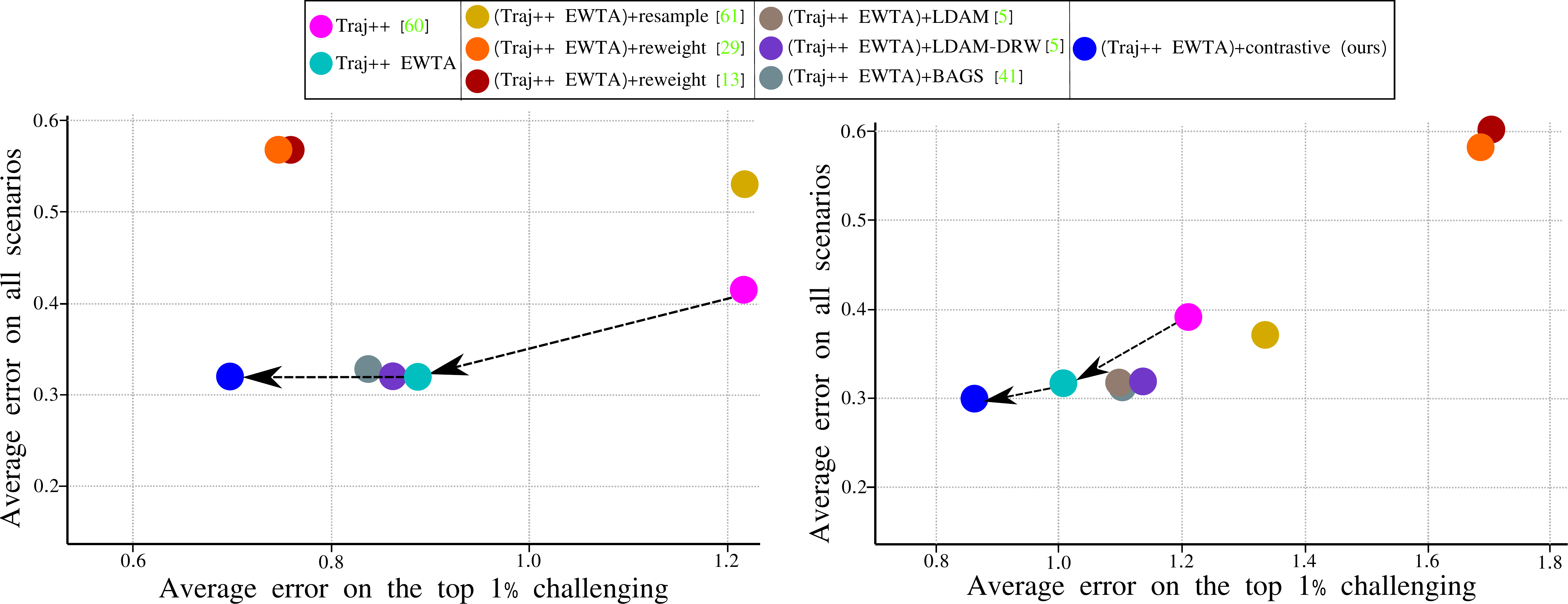}
\end{center}
\vspace*{-2mm}
  \caption{Average vs. Top 1\% error comparison on the \textbf{ETH-UCY dataset} (left) and the \textbf{nuScenes bird's eye view} (right). Our base method of integrating EWTA with the backbone of Trajectron++ (cyan) outperforms the previous state-of-the-art (magenta). Joint learning with the contrastive loss (blue) yields large improvements on the challenging scenarios while not reducing the overall average accuracy. The improvements are indicated by dashed arrows. While the resampling/reweighting baselines also improve on the hard cases, they increase the average error a lot (overfitting). The model-based baselines for long-tailed (LDAM and BAGS) yield only small improvements on ETH-UCY or worse performance on nuScenes bird's eye view.}
  \vspace*{-2mm}
\label{fig:top_plot_tab}
\end{figure*}

\begin{figure*}
\begin{center}
\includegraphics[width=1.0\textwidth]{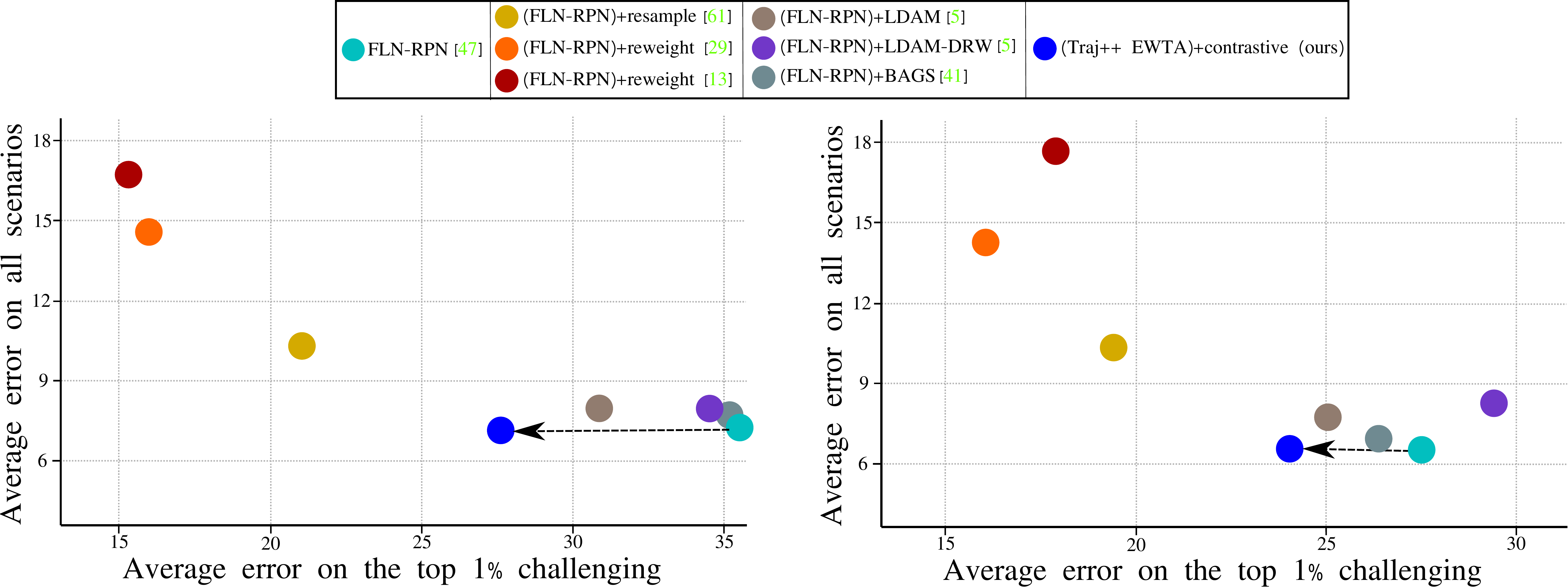}
\end{center}
\vspace*{-2mm}
  \caption{Average vs. Top 1\% error comparison on the \textbf{nuScenes egocentric view dataset} (left) and the \textbf{Waymo open dataset} (right). Our approach utilizing the contrastive loss (blue) yields a significant improvement on the challenging scenarios while not reducing the overall average accuracy. The improvements are indicated by dashed arrows. While the resampling/reweighting baselines also improve on the hard cases, they increase the average error a lot (overfitting). The model-based baselines for long-tailed (LDAM and BAGS) yield smaller improvements than our method.}
  \vspace*{-2mm}
\label{fig:ego_plot_tab}
\end{figure*}

Figure \ref{fig:top_plot_tab} and \ref{fig:ego_plot_tab} show the comparison between our method and different baselines where each circle indicates the performance of one method. These figures illustrate better the improvements gained by our method (dashed arrows).

\section{Feature Space Visualization}
\begin{figure*}[t!]
\includegraphics[width=1.0\linewidth]{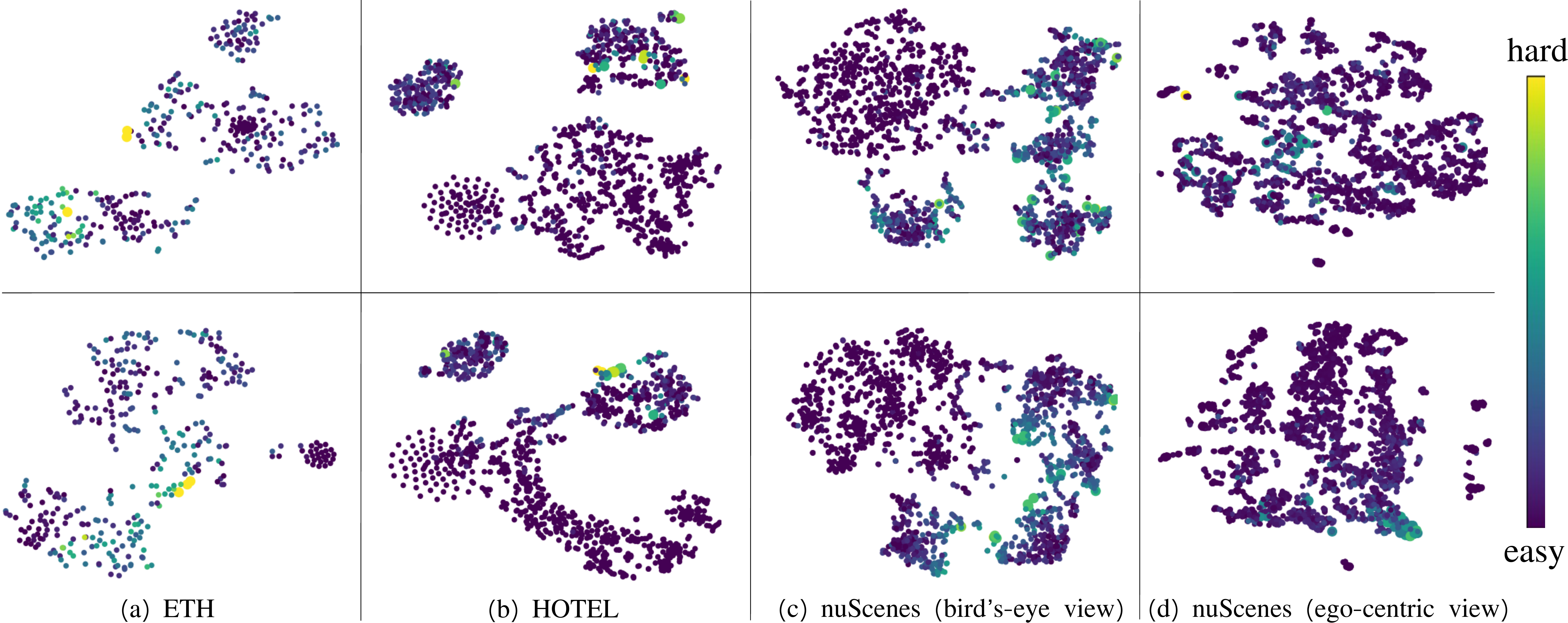}
  \caption{Plot of the feature space using tSNE \cite{tsne} on three different datasets (a and b are different scenes from the ETH-UCY dataset). \textbf{Top.} Training only with the supervised regression loss. \textbf{Bottom.} The resulting feature space when trained jointly with the contrastive loss. Large brighte circles indicate the top 1\% challenging scenarios. The darker the color of the sample, the easier it is.}
\label{fig:feats}
\end{figure*}

Figure \ref{fig:feats} shows the projection of the feature space using tSNE \cite{tsne} on three different datasets with different input modalities and views. For each dataset, we show the feature space embedding without our joint optimization (i.e, only the supervised loss) and with our joint optimization (i.e, additionally utilizing the contrastive loss). Note how our approach reshapes the feature space by pushing the challenging scenarios to be closer so that they can benefit each other as also shown in our quantitative results.

\section{Effect of the Strength of the Contrastive Loss}
In Table \ref{tab:lambda_eth} we show a study for the importance of the contrastive loss ($\lambda$) used in our approach (Eq. (4)). Using a small factor leads to small improvements on the challenging scenarios as the force of reshaping the feature space is rather weak. On the other hand, using a very large factor yields worse results as the network focuses more on reshaping the feature space and ignores the important cues for the actual task which are learned from the supervised loss. Note that this study is used only to show the effect of the weight of the contrastive loss. In our main results, we use the validation set to select the best value for $\lambda$.

\section{More Qualitative Results}
\begin{figure*}[t!]
\includegraphics[width=1.0\linewidth]{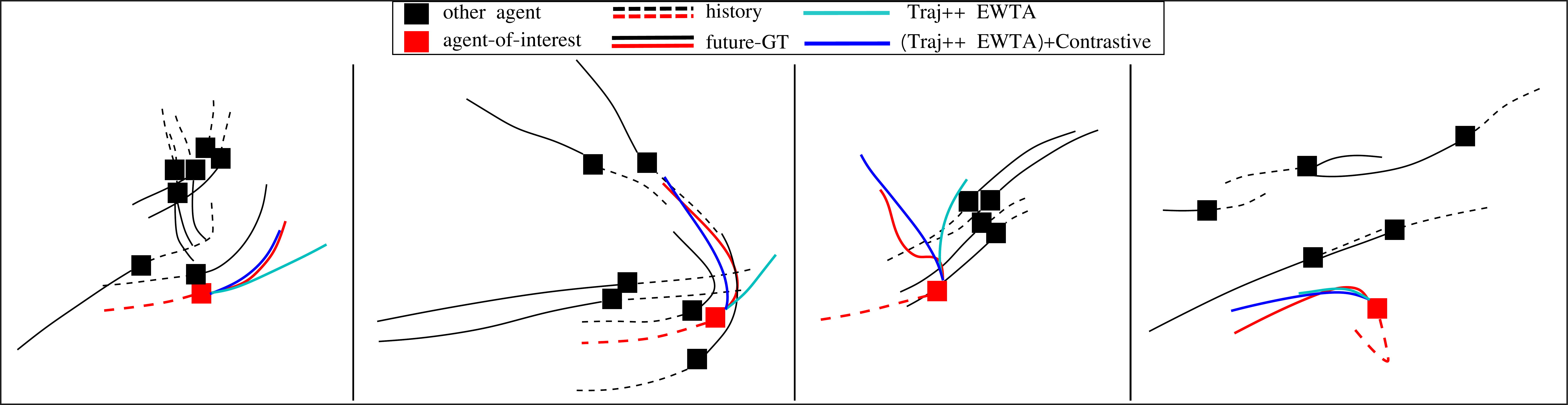}
  \caption{More results from our approach on the ETH-UCY dataset. For all these challenging scenarios, \textcolor{blue}{our approach} reasons successfully about the social relations to other pedestrians and yields better prediction than the baseline.}
\label{fig:qual}
\end{figure*}

\begin{figure*}[t!]
\includegraphics[width=1.0\linewidth]{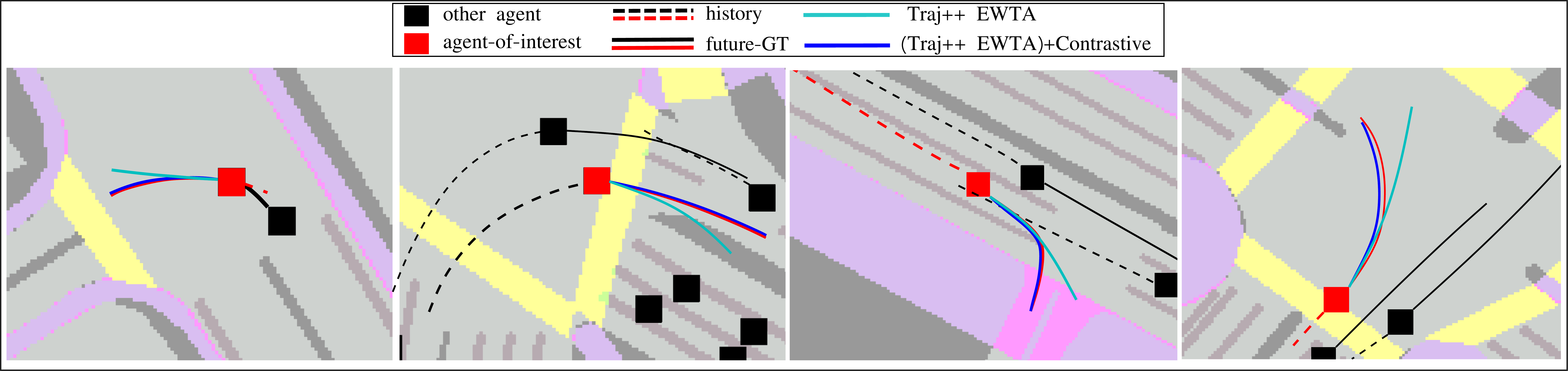}
  \caption{More results from our approach on the nuScenes dataset (bird's-eye view). For all these challenging scenarios, \textcolor{blue}{our approach} reasons successfully about the semantic cues and predicts the correct trajectory.}
\label{fig:qual2}
\end{figure*}

\begin{figure*}[t!]
\includegraphics[width=1.0\linewidth]{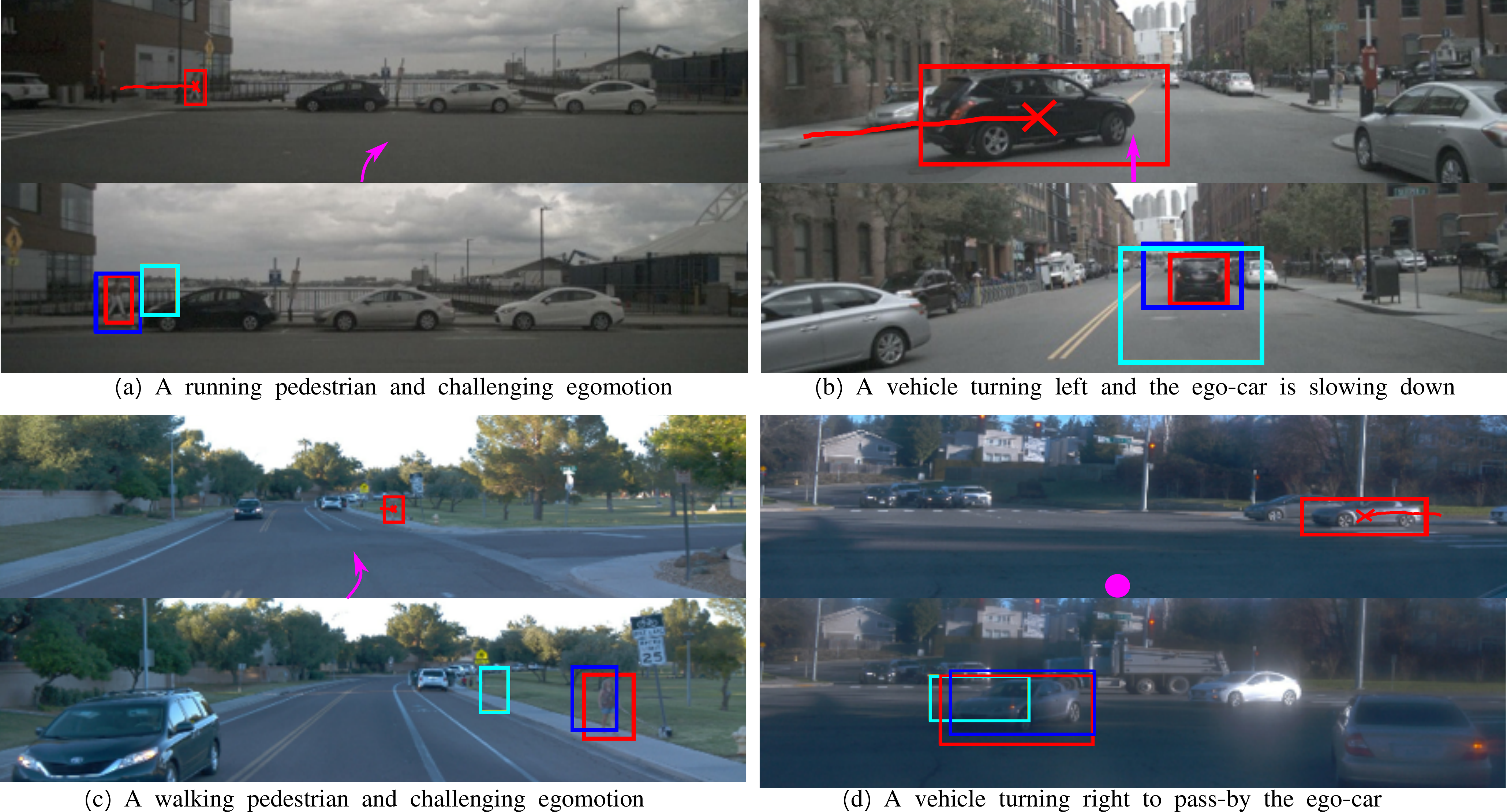}
  \caption{More results from our approach on both egocentric view datasets: nuScenes (a-b) and Waymo (c-d). For each example, we show both the last observed image (top) and the future image (bottom) along with the predictions (\textcolor{cyan}{FLN-RPN \cite{fln}} and \textcolor{blue}{Ours}) and the \textcolor{red}{ground truth}. We visualize the best hypothesis for each method. The \textcolor{magenta}{future egomotion} is also shown as arrow indicating the motion of the ego-car.}
\label{fig:qual3}
\end{figure*}

We provide more qualitative results from our approach in Figure~\ref{fig:qual}, Figure~\ref{fig:qual2} and Figure~\ref{fig:qual3} for the ETH-UYC, nuScenes (bird's-eye view) and nuScenes/Waymo (egocentric view) datasets, respectively.

\begin{table}[t]
\centering
\resizebox{1.0\linewidth}{!}{%
\begin{tabular}{|l|c|c|c|c|}%
\hline
                                                    &  \multicolumn{4}{c|}{ETH-UCY (AVG)}\\
\hline
                                                    & All                & Top 3\%   & Top 2\%   & Top 1\%   \\
\hline
Traj++ EWTA (ours)                                  & \textbf{0.16/0.32} & 0.47/1.07 & 0.51/1.13 & 0.42/0.87 \\
\hline
+ contrastive ($\lambda = 20$)                        & 0.17/0.33          & 0.47/1.04 & 0.50/1.07 & 0.43/0.84 \\
+ contrastive ($\lambda = 50$)                        & \textbf{0.16/0.32} & \textbf{0.46/1.03} & \textbf{0.48/1.03} & \textbf{0.38/0.71} \\
+ contrastive ($\lambda = 100$)                       & 0.17/0.32          & 0.48/1.04 & 0.52/1.10 & 0.50/0.97 \\
\hline
\end{tabular}
}
\vspace*{0mm}
    \caption{Study of the hyper-parameter $\lambda$ on the ETH-UCY dataset. While small $\lambda$ yields small improvement on the challenging scenarios, large $\lambda$ yields larger errors on the challenging scenarios.}
    \label{tab:lambda_eth} 
    \vspace*{-1mm}
\end{table}

\section{Detailed Quantitative Results}
Table \ref{tab:overfitting-supp} show a detailed comparison between our method and the resampling/reweighting baselines across all datasets on all metrics and difficulties. This support our findings that these baselines tend to bias the challenging cases (overfitting) while our approach maintain the average performance and improves largely on the challenging cases.

\begin{table*}[hb!]
\centering
\resizebox{1.0\linewidth}{!}{%
\begin{tabular}{|l|c|c|c|c|c|c|c|c|c|c|c|c|c|c|c|c|}%
\hline
                                              & \multicolumn{4}{c|}{ETH-UCY}                  & \multicolumn{4}{c|}{nuScenes-Bird's Eye View} & \multicolumn{4}{c|}{nuScenes Egocentric View}  & \multicolumn{4}{c|}{Waymo Open Dataset} \\
                                              & All       & Top 3\%    & Top 2\%  & Top 1\%   & All       & Top 3\%   & Top 2\%   & Top 1\%   & All   & Top 3\% & Top 2\% & Top 1\%            & All   & Top 3\% & Top 2\% & Top 1\% \\
\hline
Baseline                                      & 0.16/0.32 & 0.47/1.07 & 0.51/1.13 & 0.42/0.87 & 0.19/0.32 & 0.48/0.88 & 0.50/0.88 & 0.59/1.02 & 7.10  & 29.98   & 31.13   & 36.16              & 6.39  & 24.87   & 25.49   & 27.32    \\
\hline
+ resample~\cite{oversample0}                 & 0.25/0.53 & 0.56/1.16 & 0.61/1.24 & 0.61/1.22 & 0.21/0.37 & 0.55/0.98 & 0.61/1.07 & 0.78/1.33 & 10.20 & 18.90   & 19.37   & 21.62              & 10.48 & 19.46   & 18.91   & 19.69    \\
+ reweight~\cite{reweight1-inverse-freq}      & 0.28/0.56 & \textbf{0.41/0.78} & \textbf{0.44/0.81} & 0.43/0.76 & 0.33/0.58 & 0.74/1.28 & 0.80/1.38 & 0.99/1.67 & 14.47 & 15.33   & 15.42   & 16.20              & 14.00 & \textbf{17.01}   & \textbf{16.80}   & \textbf{16.44}    \\
+ reweight~\cite{reweight4-inverse-effective} & 0.28/0.56 & 0.43/0.83 & 0.45/0.86 & 0.44/0.78 & 0.34/0.60 & 0.75/1.33 & 0.80/1.42 & 0.99/1.71 & 16.54 & \textbf{15.29}   & \textbf{15.34}   & \textbf{15.46}              & 17.43 & 20.34   & 19.40   & 18.79    \\
\hline
+ contrastive                                 & \textbf{0.16/0.32} & 0.46/1.03 & 0.48/1.03 & \textbf{0.38/0.71} & \textbf{0.18/0.30} & \textbf{0.44/0.73} & \textbf{0.46/0.72} & \textbf{0.54/0.85} & \textbf{7.04}  & 25.05   & 25.26   & 27.49              & 6.49  & 22.36   & 22.72   & 24.09    \\
\hline
\end{tabular}
}
\vspace*{0mm}
    \caption{Comparison to the common resampling/reweighting techniques on the four datasets. For each method, we show the min-FDE/min-ADE over all samples and over top 1-3\% challenging samples. Our method yields large improvements on the challenging ones while maintaining the average. This is in contrast to the reweighting/resampling baselines, which lead to much worse performance on average (see the error increase on the 'All' columns). Baseline indicates Traj++ EWTA for bird's eye view and FLN-RPN \cite{fln} for egocentric view.}
    \label{tab:overfitting-supp} 
    \vspace*{-3mm}
\end{table*}

\section{Baselines Implementation Details}
In order to use state-of-the-art methods for long-tail classification, we map the regression task to a classification task by assigning classes to training samples based on the error of the Kalman filter. In particular, we group the errors into bins and assign the same class to all samples in each bin. To alleviate the issue of having classes with only one sample, we group all samples with a score greater than a specific threshold into the same bin. This yields 13, 36, 331 classes for ETH-UCY, nuScenes bird's eye view and nuScenes egocentric view, respectively. For all baselines (including our method), we use the same joint training scheme where two heads (classification and regression) are trained on top of the feature embedding. For the LDAM baseline~\cite{ldam-loss}, we experiment with different scaling factors and use the best setting $s=1$. Following BAGS~\cite{softmax-balanced}, we split the classes into 4 homogeneous groups to ensure that all classes from the same group have roughly the same number of items and use a sampling ration of 8 to ensure that all groups contribute to the mini-batch during training.



\end{document}